\documentclass[sigconf]{acmart}

\usepackage{epsfig}
\usepackage{graphicx}
\usepackage{amsmath}
\usepackage{engord}
\usepackage{bbding}
\usepackage{adjustbox}
\usepackage{verbatim}
\usepackage{bm}
\usepackage[ruled]{algorithm2e}
\usepackage{color}
\usepackage{multirow}

\usepackage{comment}
\usepackage{tabu}
\usepackage{arydshln}

\usepackage{balance}

\AtBeginDocument{%
  \providecommand\BibTeX{{%
    \normalfont B\kern-0.5em{\scshape i\kern-0.25em b}\kern-0.8em\TeX}}}

\copyrightyear{2021}
\acmYear{2021}
\setcopyright{acmcopyright}\acmConference[MM '21]{Proceedings of the 29th ACM
International Conference on Multimedia}{October 20--24, 2021}{Virtual Event, China}
\acmBooktitle{Proceedings of the 29th ACM International Conference on Multimedia
(MM '21), October 20--24, 2021, Virtual Event, China}
\acmPrice{15.00}
\acmDOI{10.1145/3474085.3475217}
\acmISBN{978-1-4503-8651-7/21/10}

\begin{document}
\fancyhead{}

\title{WeClick: Weakly-Supervised Video Semantic Segmentation \\ with Click Annotations}

\author{
Peidong Liu{\normalsize$^{1, \star}$}, Zibin He{\normalsize$^{1, \star}$}, Xiyu Yan{\normalsize$^{1, \star}$}, Yong Jiang{\normalsize$^{1,2}$}, Shu-Tao Xia{\normalsize$^{1,2, \dagger}$} \\ Feng Zheng{\normalsize$^{3}$}, Maowei Hu{\normalsize$^{1,4}$}
}

\affiliation{
    \country{$^1$Tsinghua Shenzhen International Graduate School, Tsinghua University} \\
    \country{$^2$PCL Research Center of Networks and Communications, Peng Cheng Laboratory} \\
    \country{$^3$Department of Computer Science and Engineering,
    Southern University of Science and Technology} \\
    \country{$^4$Shenzhen Rejoice Sport Tech. Co., LTD}
}
\thanks{$\dagger$ Corresponding author.}
\thanks{$\star$ Equally contributed.}
\thanks{This work is supported in part by the National Key Research and Development Program of China under Grant 2018YFB1800204 and the National Natural Science Foundation of China under Grant 61771273, 61972188.}

\email{{lpd19,hzb19,yanqy17}mails.tsinghua.edu.cn, {jiangy,xiast}@sz.tsinghua.edu.cn, zhengf@sustech.edu.cn, humaowei@51yund.com}


\begin{abstract}
Compared with tedious per-pixel mask annotating, it is much easier to annotate data by clicks, which costs only several seconds for an image. However, applying clicks to learn video semantic segmentation model has not been explored before. In this work, we propose an effective weakly-supervised video semantic segmentation pipeline with click annotations, called \textit{WeClick}, for saving laborious annotating effort by segmenting an instance of the semantic class with only a single click. Since detailed semantic information is not captured by clicks, directly training with click labels leads to poor segmentation predictions. To mitigate this problem, we design a novel \textit{memory flow knowledge distillation} strategy to exploit temporal information (named \textit{memory flow}) in abundant unlabeled video frames, by distilling the neighboring predictions to the target frame via estimated motion. Moreover, we adopt vanilla knowledge distillation for model compression. In this case, WeClick learns compact video semantic segmentation models with the low-cost click annotations during the training phase yet achieves real-time and accurate models during the inference period. Experimental results on Cityscapes and Camvid show that WeClick outperforms the state-of-the-art methods, increases performance by 10.24\% mIoU than baseline, and achieves real-time execution.

\end{abstract}




\maketitle

\section{Introduction}  

\begin{figure}[t]
  \newlength\fsdurthree
  \centering
  \begin{tabular}{cc}
    \centering
    \hspace{-4mm}
    \begin{adjustbox}{valign=t}
    \small
      \begin{tabular}{c}
        \includegraphics[width=4cm,height=2.2cm]{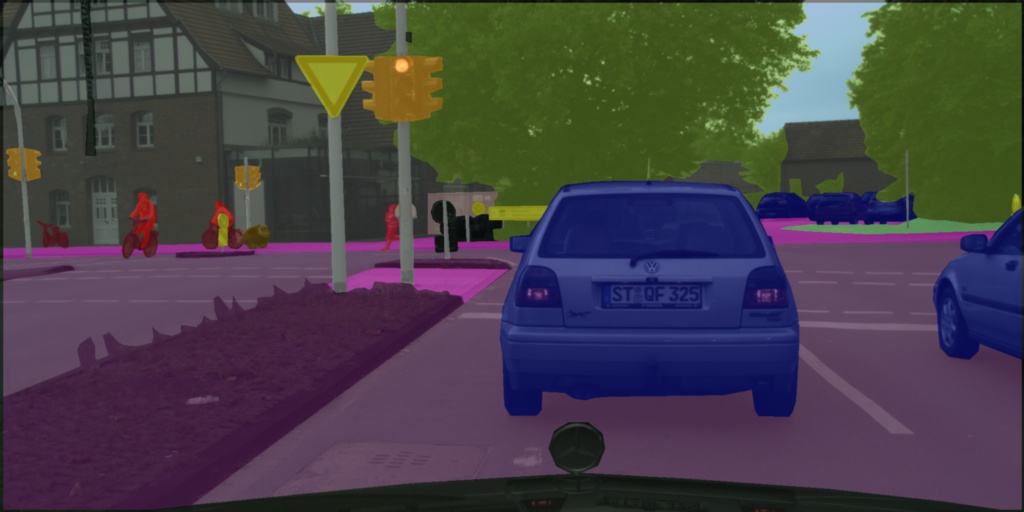}  
        \\ 
        (a) Fine annotations. 
        \\
        \includegraphics[width=4cm,height=2.2cm]{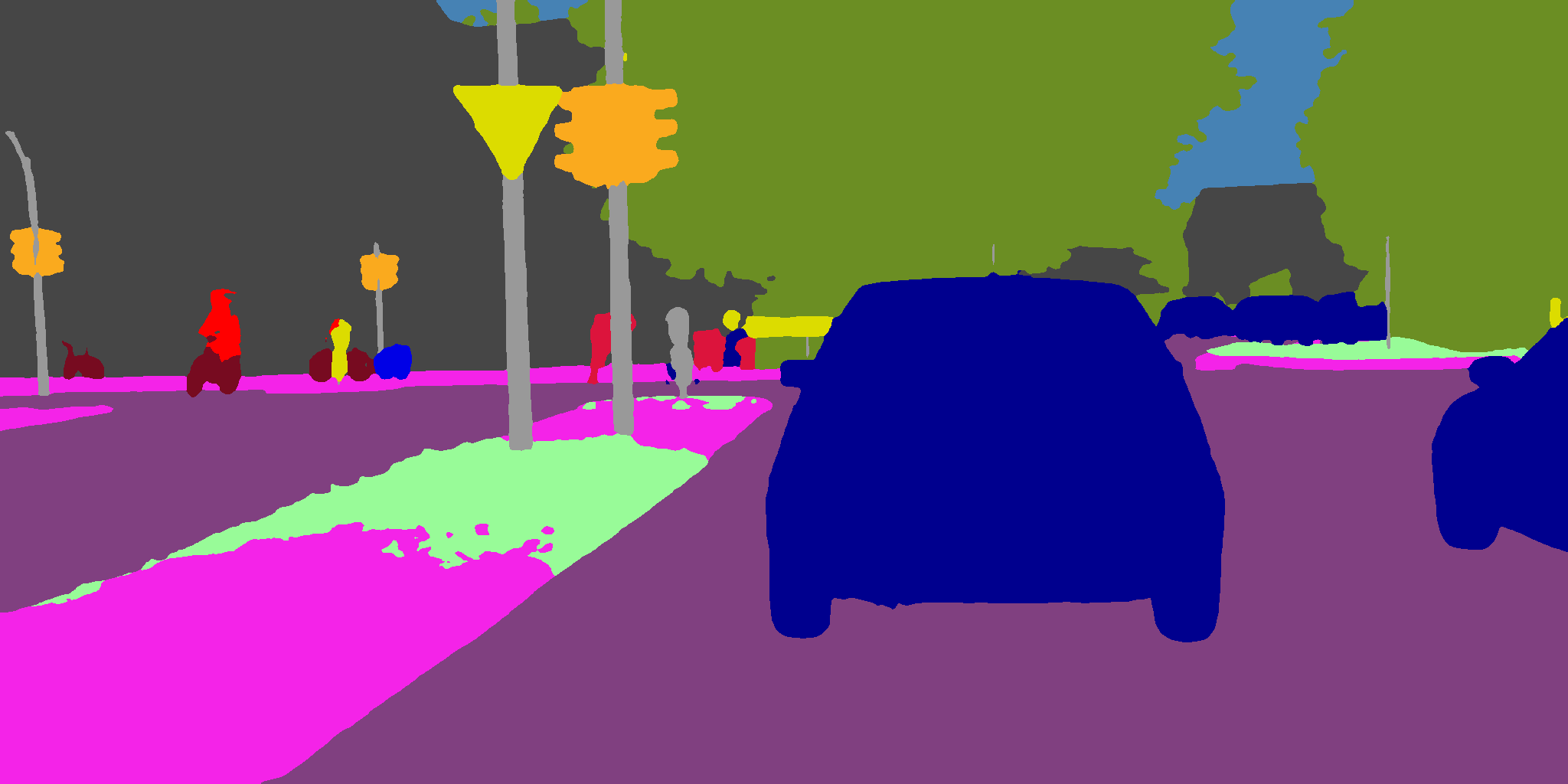}  
        \\ 
        (c) Result with fine annotations.
        \\
        
      \end{tabular}
    \end{adjustbox}
    
    \hspace{-4mm}

    \begin{adjustbox}{valign=t}
    \small
      \begin{tabular}{c}
        \includegraphics[width=4cm,height=2.2cm]{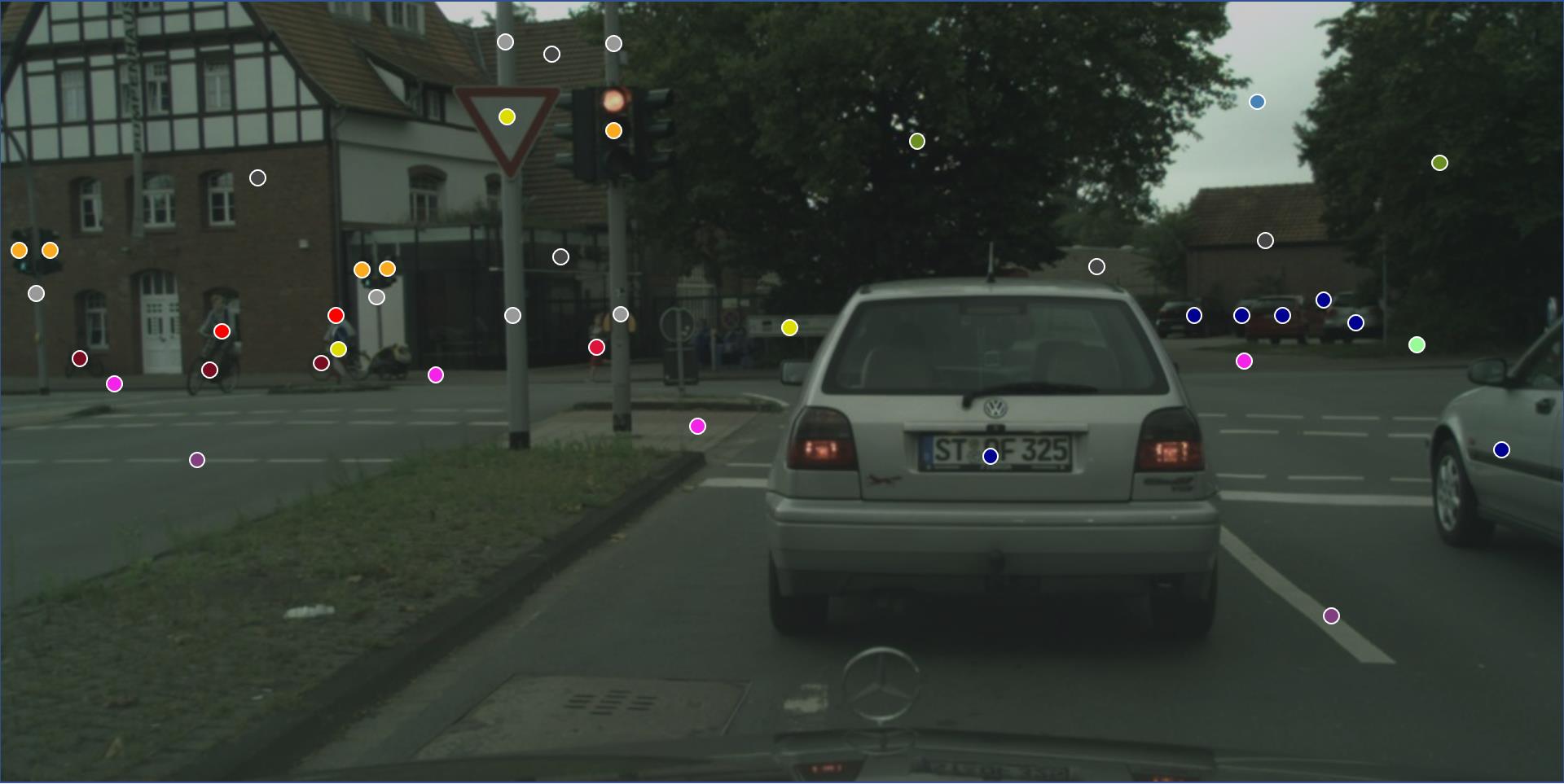}  
        \\ 
        (b) Click annotations. 
        \\
        \includegraphics[width=4cm,height=2.2cm]{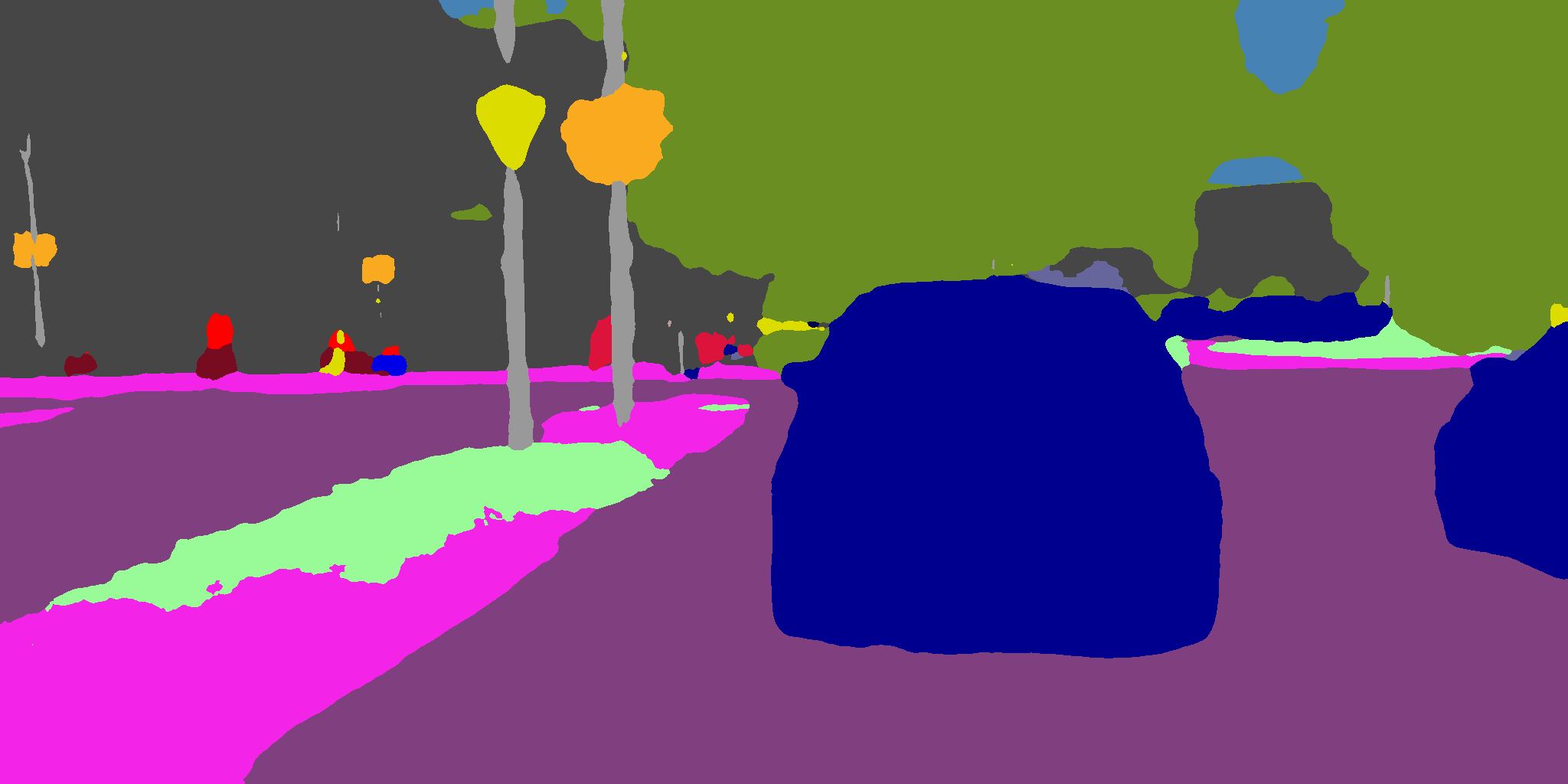}  
        \\ 
        (d) Result with click annotations.  
        \\
      \end{tabular}
    \end{adjustbox}

  \end{tabular}  

  \caption{
    Examples of (a) fine annotations and (b) click annotations for an image from Cityscapes validation set. (c) and (d) are the semantic segmentation results trained with fine annotations and click annotations, respectively. It can be observed that (d) achieves comparable segmentation results with (c) in some instance class, e.g. car,  person, rider.
    }
\label{fig:first-fig}
\end{figure}


Video semantic segmentation (VSS), one of the fundamental high-level tasks in computer vision, aims to assign a semantic label to each pixel in every video frame. Compared with image semantic segmentation, VSS has a wider range of application scenarios, e.g. autonomous driving and human-computer interaction~\cite{yan2020video}, which deserves more attention from researchers. 

In recent years, the community has witnessed substantial progress in VSS. However, VSS still encounters major bottlenecks because most existing VSS methods~\cite{stgru, li2018low, netwarp, dff} depend on large-scale per-pixel masks to learn sufficient semantic information for obtaining well-performed models. Annotating per-pixel masks is time-consuming because it takes an annotator 1.5 hours on average to label a frame. Therefore, Cityscape dataset~\cite{Cordts2016} annotates only 1 frame for a 30-frame video snippet to reduce huge human effort.

To cope with the fine-annotation scarcity issue, a natural way is to use weakly-supervised annotations to train VSS networks. Researchers have explored several weakly-supervised methods in image semantic segmentation, including click-based~\cite{bearman2016whats,obukhov2019gated},  box-based~\cite{dai2015boxsup,khoreva2017simple,ibrahim2018weakly}, and extreme point-based~\cite{maninis2018deep, wang2019object} approaches. Among these weakly-supervised approaches, click-based methods reduce the tedious labeling process to the greatest extent because it costs the annotator only several seconds for an image. Figure~\ref{fig:first-fig}(a) and (b) shows the difference between per-pixel masks and click annotations. However, to the best of our knowledge, little attention has been paid to the weakly-supervised learning algorithms with click annotations in VSS. In order to promote the development of this field, we extend the classical click-based learning method~\cite{obukhov2019gated} in image semantic segmentation to video semantic segmentation for learning semantic information with low-cost click labels in video data, which is named the \textit{weakly} training scheme in this work.

\begin{figure}[t]
\center
\vspace{-1.5mm}
\includegraphics[width=1.0\linewidth]{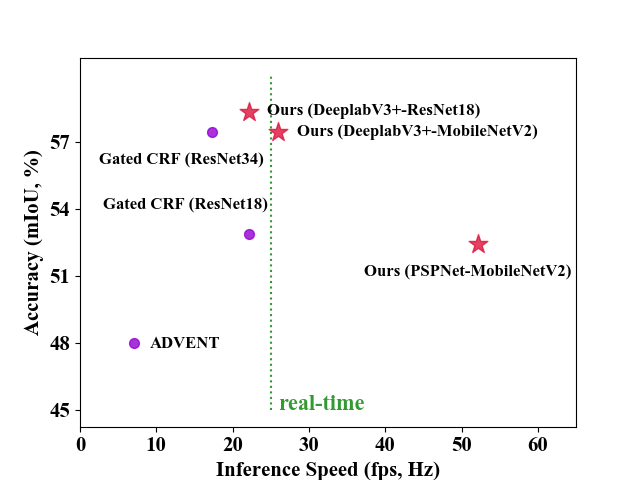}
\caption{Comparison on the accuracy and inference speed of our method and the state-of-the-art weakly-supervised methods: Gated CRF~\cite{obukhov2019gated} (click-based image semantic segmentation) and ADVENT~\cite{Vu2019ADVENT} (unsupervised domain adaptation image semantic segmentation). The experiment is conducted on Cityscapes with an Nvidia Tesla V100 GPU.}
\label{fig:fps}
\end{figure}
However, directly training with click annotations leads to poor segmentation performance because detailed semantic information is not captured by clicks. Inspired by the previous self-distillation work~\cite{zhang2019your, hou2019learning}, to properly exploit temporal information in plenty of unlabeled video frames, we design a novel temporal self-distillation mechanism, called \textit{Memory Flow knowledge Distillation (MFD)} training scheme, which obtains internal temporal information (named \textit{memory flow}) within the network itself by distilling the neighboring predictions as extra supervision signals to the target frame. Specifically, a pre-trained optical flow net is applied to predict the optical flow of each pixel from the neighboring frame to the target frame based on the frame-pair input. Then the neighboring predictions are warped to the target frame with the estimated optical flow to supervise the learning of the target frame. Besides, a well-designed consistency matrix is employed to regularize the propagation of the estimated motion by adaptively tuning the importance of memory flow in the MFD training scheme.

In this paper, we propose an effective weakly-supervised video semantic segmentation pipeline with click annotations, called \textit{WeClick}, which learns models with low-cost click labels in the training phase yet achieves fast and accurate prediction without any additional computation cost and post-processing during inference. In specific, WeClick applies the weakly training scheme to alleviate the issue of fine-annotation scarcity and the MFD training scheme to exploit the temporal knowledge. Moreover, to further ease the inference latency problem, WeClick adopts vanilla knowledge distillation for model compression. After training the compact model, all the teacher net and the motion estimation net in the training phase are removed. Benefiting from the proposed training schemes, as Figure~\ref{fig:first-fig}(c) and (d) shows, WeClick achieves comparable segmentation results with the models trained with fine annotations. During inference, WeClick only keeps the student net as the VSS model with single-frame input so that the proposed training schemes boost the model accuracy without additional computation cost. As Figure~\ref{fig:fps} illustrates, our DeeplabV3+ with MobileNetV2 can reach higher accuracy (i.e. mIoU) with a faster inference speed compared with other state-of-the-art weakly-supervised methods.

Extensive experiments on benchmark Cityscapes and Camvid demonstrate the effectiveness and generalization of our method. WeClick outperforms the state-of-the-art methods based on image-level and click-level annotations and the most significant improvement in terms of mIoU reaches 10.24\% than baseline on Cityscapes. Several lightweight backbones including ResNet18, ResNet34~\cite{he2016deep} and MobileNetV2~\cite{mobilenetv2}, empirically verify that WeClick improves segmentation accuracy with a faster inference speed. 


In summary, our main contributions are as follows:
\begin{itemize}
  \item 
  We propose the \textit{WeClick} pipeline, the first study to explore click-based learning in VSS, which learns compact models with low-cost click labels in the training phase yet achieves real-time and accurate prediction in the inference period.
  \item 
  We employ the \textit{weakly} learning scheme, which is extended from the click-based method in image semantic segmentation, to mitigate the fine-annotation scarcity issue in VSS. 
  \item 
  To boost the segmentation performance, we design a novel \textit{Memory Flow knowledge Distillation (MFD)} scheme, which utilizes memory flow in plentiful unlabeled video frames.
  \item 
  Empirical experiments on Cityscapes and Camvid show that the compact models with the proposed training schemes outperform other state-of-the-art weakly-supervised VSS methods substantially.

\end{itemize}

\section{Related Work}  
\subsection{Video Semantic Segmentation}
Video semantic segmentation (VSS), also known as video scene parsing (VSP), refers to the process of allotting a semantic label for each pixel in every video frame~\cite{yan2020video}. 
VSS is greatly different from image semantic segmentation in that the former is characterized by abundant temporal information. According to the usage of optical flow~\cite{dff, tracking}, previous VSS works are summarized into two categories. The first one is to accelerate VSS algorithms by propagating the predictions of previous frames to the target frame using optical flow. For example, the methods~\cite{dff,ilg2017flownet,xu2018dynamic,li2018low} obtains the segmentation result of the next frame by processing the feature map or segmentation mask of the previous frame via optical flow, thus greatly reducing redundant calculations in the video segmentation. However, the accuracy of the segmentation is reduced. The second one is to use optical flow and other modules to fuse the features of the preceding and subsequent frames or add constraints to learn stronger representation ability for higher accuracy of single frame semantic segmentation~\cite{
netwarp}. Our work belongs to the second category with weakly-supervised annotations.

\subsection{Weakly-Supervised Segmentation}
In recent years, various types of weakly-supervised semantic segmentation techniques have been studied and developed to reduce the demand for large-scale detailed annotations. 
The typical weakly-supervised forms of image semantic segmentation are image category tagging~\cite{wei2016learning,sadat2017bringing,wang2020deep,wang2020weakly}, bounding box annotations~\cite{dai2015boxsup,khoreva2017simple,ibrahim2018weakly}, scribble annotations~\cite{lin2016scribblesup,cciccek20163d}, point annotations~\cite{bearman2016whats,obukhov2019gated} and eye movement annotations~\cite{Papadopoulos2014training}. We focus on the weakly-supervised learning methods based on the low-cost click annotations.

In this context, a series of weakly-supervised learning methods based on point labeling are investigated. Firstly, inspired by a human-oriented object, Bearman \emph{etal}~\cite{bearman2016whats} proposes a supervision mechanism based on point label in each class for training networks. 
Besides, they prove that the model of point-level supervision training is better than that of the image-level supervision with a fixed annotating budget by manually annotating and evaluating the annotation time on the Pascal VOC 2012 dataset~\cite{everingham2010pascal,hariharan2011semantic}. The above work is the beginning of point labeling supervision. Tang \emph{etal}~\cite{tang2018normalized} studies a new method on this basis, in order to minimize the performance gap between weakly- and fully-supervised semantic segmentation. They first propose to jointly minimize a partial cross-entropy loss and a regularized loss for labeled pixels and unlabeled pixels respectively. Next, they further extend the regularization loss~\cite{tang2018regularized} to a more general loss function, such as graph cuts or dense CRFs. The aforementioned methods somewhat rely on alternative sources of supervision such as pre-training on other datasets. Recently, Obukhov \emph{etal}~\cite{obukhov2019gated} propose Gated CRF loss for the unlabeled pixels. It can be trained easily with the standard Stochastic Gradient Descent algorithm without any pre- and post-processing operations and achieves state-of-the-art performance for click-based image semantic segmentation algorithms. However, the VSS models, which are directly trained with Gated CRF loss, perform poorly for limited semantic information in click annotations. Therefore, we extend Gated CRF loss to the VSS model for click-based learning and further utilize temporal information in plenty of unlabeled video data via the proposed training schemes.


The above methods focus on image semantic segmentation, which is based on the relatively simple images in the Pascal VOC dataset. In fact, VSS in complex scenes has a more expensive annotation cost and needs to be broken through in weakly-supervised methods~\cite{2020Weakly,chen2020semi,netwarp}. For examples, Saleh \emph{etal} proposes a weakly-supervised two-stream (WSTS) method for VSP \cite{Saleh2017}, which handles foreground and background objects evenly, with one stream taking the image and the other taking the optical flow to extract features. \cite{Lee2019ICCV} proposes a weakly-supervised method that uses temporal information to train a network on a video dataset labeled at the image level, which is automatically harvested from the web. The model obtains activated regions from each video frame and then aggregates them in a single image. In the work of \cite{chen2020semi}, they use unlabeled video sequences to improve the image semantic segmentation or instance segmentation of urban scenes segmentation. Instead, our method concentrates on improving the performance of weakly-supervised VSS methods with clicks in more complex scenes~\cite{Cordts2016}.

\subsection{Knowledge Distillation}


Besides model pruning~\cite{han2015deep, liu2017learning, he2017channel}, weight quantization~\cite{han2015deep, xnor} and compact network design~\cite{mobilenets, mobilenetv2}, knowledge distillation~\cite{hinton2015distilling} is first proposed as a novel technique for network compression. It is characterized by a Teacher-Student learning paradigm that propagates ``dark knowledge'' of a cumbersome teacher net to supervise the training of a tiny student net. Specifically, knowledge includes soft labels~\cite{hinton2015distilling}, intermediate features~\cite{fitnet, AT, FSP, ft, AB}, correlation information~\cite{SP, CC, IRG, RKD}, etc.

Although researches on knowledge distillation mostly focus on the image classification task, it can also be extended to more complex visual tasks, including object detection~\cite{chen2017learning, li2017mimicking}, pose estimation~\cite{fastpose} and image restoration~\cite{lee2020learning, FAKD}, which reveals the generality of this learning framework. The pioneer knowledge distillation work in semantic segmentation~\cite{xie2018improving} transfers pixel-wise class probabilities and segmentation boundaries to the student net. Liu \emph{etal}~\cite{liu2019structured} proposes structured distillation, including pair-wise distillation that extracts feature similarity in a local patch, and holistic distillation which captures higher-order semantics. Previous work in VSS that is related to our work is~\cite{liu2020efficient}, which preserves temporal consistency by encoding motion information in distillation loss terms. In contrast, we concentrate on weakly-supervised learning with click annotations in VSS. 

Self-distillation, also named teacher-free distillation, is proposed to prevent the usage of large teacher nets, which distills knowledge within the network itself. The commonly-used self-distillation strategy includes transferring high-level features to low-level features~\cite{zhang2019your, hou2019learning}, using soft labels of previous epochs to guide the training of the target epoch~\cite{yang2019snapshot}, etc. In this work, we design a temporal self-distillation mechanism, i.e. memory flow knowledge distillation training scheme, to boost the segmentation accuracy.

\begin{figure*}[t]
\center
\vspace{-5mm}
\includegraphics[width=0.8\linewidth]{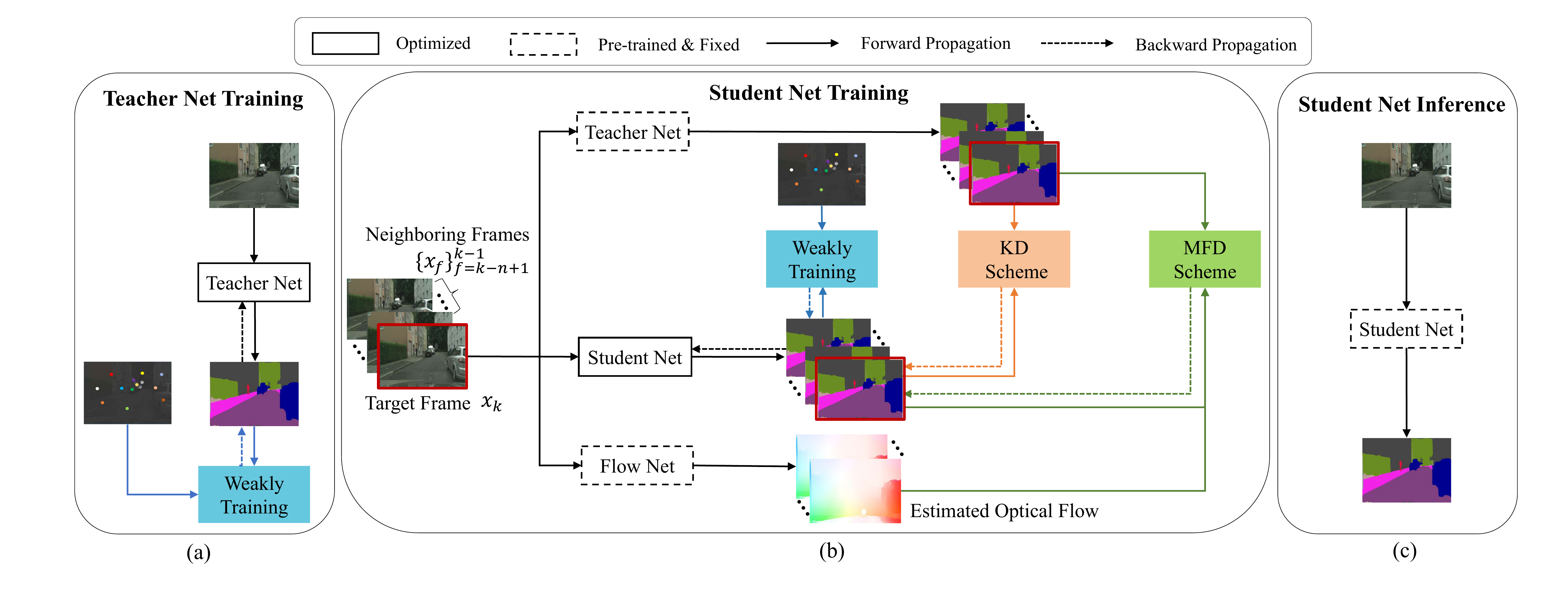}
\caption{An overview of the proposed WeClick pipeline. (a) In the teacher net training phase, we pre-trained the cumbersome teacher net only in the weakly training scheme via click annotations to obtain a segmentation network with high accuracy. (b) During the student net training period, the teacher net and the flow net are pre-trained and their weights are fixed to assist the optimization of the student net. The input of the student net is based on multi-frames input pattern, which consists of the target frame  $\boldsymbol x_k$ (\textit{marked with red border in the Figure}) with click annotations and the $n-1$ neighboring frames $\{\boldsymbol{x}_f\}_{f=k-n+1}^{k-1}$ without any labels, where $f$ denotes the frame index. The overall optimization of student net consists of three training schemes: 1) weakly training scheme to realize weakly-supervised learning with click annotations in VSS; 2) Memory Flow knowledge Distillation (MFD) scheme that exploits memory flow in abundant unlabeled video data; 3) Knowledge Distillation (KD) scheme for model compression. (c) In the student net inference phase, the teacher net and the flow net are removed and only the student net remains with single-frame inference without additional computation cost.}

\label{fig:framework}

\end{figure*}


\section{Method}
In this section, we first introduce the preliminary weakly-supervised learning with click annotations in Section~\ref{sec:Preliminary}. Then we present the proposed WeClick pipeline, including the training and inference phases in Section~\ref{sec:WeClick}. Finally we elaborate on the details of the overall training schemes in Section~\ref{ourlosses}.



\subsection{Preliminary Weakly-Supervised Learning with Click Annotations}
\label{sec:Preliminary}
Training an image semantic segmentation network requires pairs of input image $\boldsymbol{x}$ and its annotation $\boldsymbol{y}$, where each pixel is assigned from $C$ classes. Let us denote $N$ as the total number of pixels in an image, $\hat{\boldsymbol y}_i(c)$ as the $c^{th}$ class probability for pixel $\boldsymbol x_i$ in the prediction $\hat{\boldsymbol y}=\mathcal{F}(\boldsymbol x)$, where $\mathcal{F}$ is a segmentation network. In the click-based learning, only a few pixels of the training images are annotated, forming a partial map $\boldsymbol m$ of the same size as $\boldsymbol{y}$, where the element $m_i \in \{0,1\}$. The sparsity of $\boldsymbol m$ depends on whether the corresponding pixel is a click annotation. Since the unannotated pixels provide little information in the learning process, they are left out of consideration during the training phase. In general, partial Cross-Entropy (pCE) loss is used in the click-based learning algorithm:
\begin{equation}
\small
\label{eq.pce}
\mathcal{L}_{pCE}(\hat{\boldsymbol y},\boldsymbol{y}) =
\cfrac{\sum_{i=1}^N \sum_{c=1}^C m_i \left[- \boldsymbol y_i(c)\log \hat{\boldsymbol y}_i(c)\right]}{\sum_{i=1}^N m_i}.
\end{equation}

In click-based learning, the point annotation is usually applied as the supervision signal seed and then propagated to the surrounding pixels through regularization mechanisms~\cite{tang2018regularized,obukhov2019gated} for obtaining more semantic information. In other words, the preliminary click-based learning algorithm consists of two parts, i.e. the pCE loss term and the regularization term. In this work, the Gated CRF~\cite{obukhov2019gated} is used as the regularization term for its great ability in mining semantic information from click annotations.

\subsection{WeClick Pipeline}
\label{sec:WeClick}

We propose the WeClick pipeline for alleviating the fine-annotation scarcity issue in video semantic segmentation with effective training schemes. An overview of WeClick shows in Figure~\ref{fig:framework}. 

In the teacher net training phase, we pre-trained the large teacher net in the weakly training scheme with only click labels to obtain a well-performed segmentation network (see Figure~\ref{fig:framework}(a)). As Figure~\ref{fig:framework}(b) illustrates, during the student net training phase, we apply the pre-trained teacher net and the pre-trained flow net and freeze their parameters to assist the optimization of the student net. The input of the student net is based on multi-frames input pattern, which consists of the target frame  $\boldsymbol x_k$ with click annotations and the $n-1$ neighboring frames  $\{\boldsymbol{x}_f\}_{f=k-n+1}^{k-1}$ without any labels, where $f$ denotes the frame index. We train the student net with three training schemes: 1) weakly training scheme, where only the target frame $\boldsymbol x_k$ is fed to the student net to learn semantic information from click annotations; 2) Memory Flow knowledge Distillation (MFD) scheme which utilizes memory flow by distilling the neighboring predictions as extra supervised signals to the target frame via estimated optical flow; 3) Knowledge Distillation (KD) scheme, which learns soft knowledge from well-performed teacher net in both the target frame $\boldsymbol x_k$ and the neighboring frames $\{\boldsymbol{x}_f\}_{f=k-n+1}^{k}$. In the inference process, the teacher net and the optical flow net are removed and only the compact student net remains for single-frame inference with no additional computation cost (see Figure~\ref{fig:framework}(c)).

\begin{figure*}[t]
\center

\vspace{-5mm}
\includegraphics[width=0.8\linewidth]{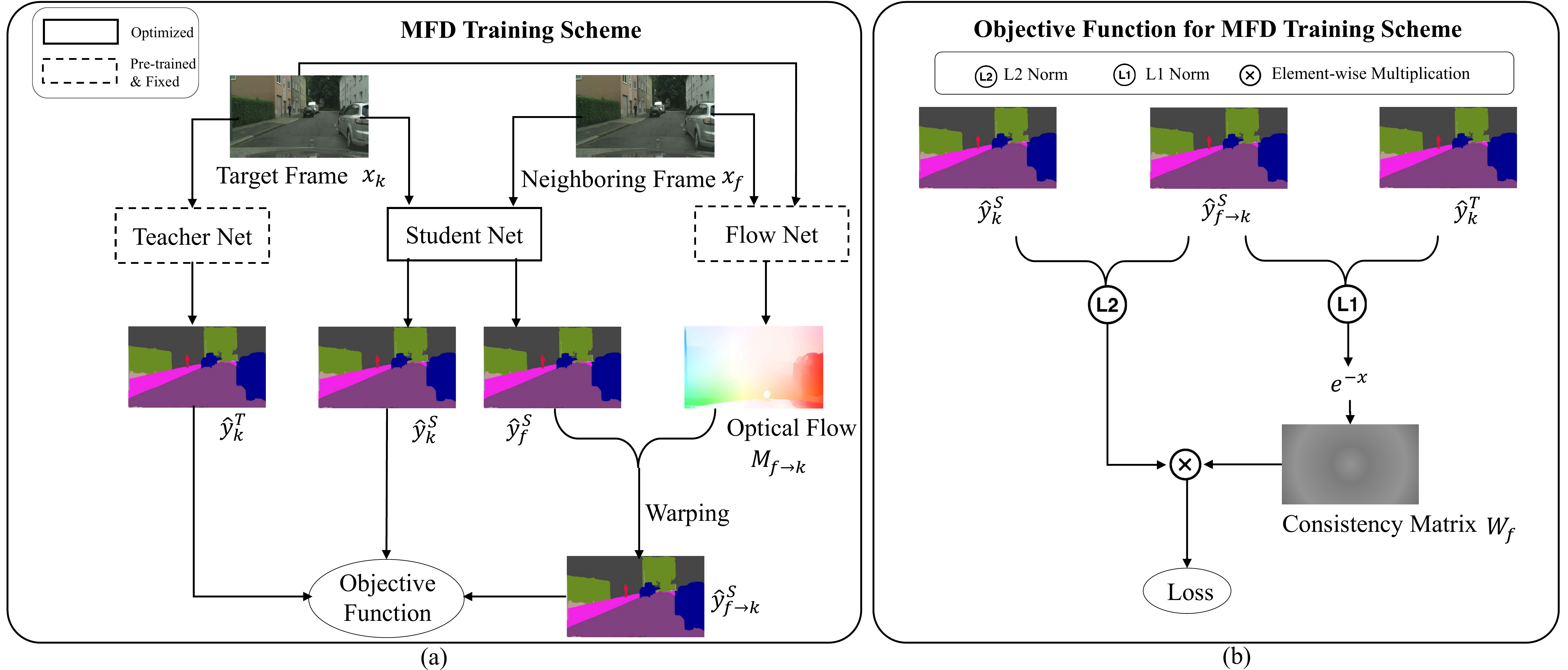}
\caption{(a) In the MFD training scheme, both the teacher net and the optical flow net are pre-trained and their parameters are fixed. Optical flow map $M_{f \rightarrow k}$ between the target frame $\boldsymbol x_k$ and the neighboring frame $\boldsymbol x_f$ is warped with the neighboring frame prediction $\hat{\boldsymbol y}_{f}^S $ to get the warped mask $\hat{\boldsymbol y}_{f\rightarrow k}^S$. Note that the teacher net is only responsible for evaluating the quality of the warped mask instead of directly providing soft labels to the student net, which is different from the common knowledge distillation. Therefore, the objective of the MFD training scheme is to minimize the gap between $\hat{\boldsymbol y}_{k}^S$ and $\hat{\boldsymbol y}_{f\rightarrow k}^S$ with the assist of the teacher net. (b) A consistency matrix $W_f$ is introduced to evaluate the quality of the warped mask $\hat{\boldsymbol y}_{f\rightarrow k}^S$. During the student net training phase, the consistency matrix adaptively decreases the weight of the unreliable region in the warped mask.}
\label{fig:MFD}
\end{figure*}

\subsection{Proposed Training Schemes}
\label{ourlosses}
In this section, we elaborate on the detailed training schemes in WeClick, which consists of three parts: weakly training scheme, MFD strategy, and KD strategy.


\noindent{\textbf{Weakly Training Scheme.}}
In WeClick, only the target frame $\boldsymbol x_k$ has click annotation $\boldsymbol y_k$ while  other neighboring frames are unannotated. We applied \textit{weakly} loss to the target frame prediction of student net $\hat{\boldsymbol y}_k^S$, which is formulated as Eq.~(\ref{eq.weak}):
\begin{equation}
\label{eq.weak}
\small
\mathcal{L}_{weakly} (\boldsymbol x_k) = \mathcal{L}_{pCE}(\hat{\boldsymbol y}_k^S, \boldsymbol{y}_k)+\lambda \mathcal{L}_{GCRF}(\hat{ \boldsymbol y}_k^S),
\end{equation}where GCRF stands for Gated CRF regularization terms~\cite{obukhov2019gated} and $\lambda$ is the loss weight hyper-parameter.


\noindent{\textbf{MFD Scheme.}} MFD is a temporal self-distillation mechanism that distills knowledge within the network itself by transferring information from neighboring frames to the target frame via the estimated optical flow. The detailed MFD training scheme is illustrated in Figure~\ref{fig:MFD}(a). 
We apply a pre-trained motion estimation (i.e. optical flow) net $g(\cdot)$ to estimate the optical flow map from the $f^{th}$ frame $\boldsymbol x_f$ to the target frame $\boldsymbol x_k$, i.e. $M_{f \rightarrow k} = g(\boldsymbol x_{f}, \boldsymbol x_{k}) \in R^{N \times 2}$, where $N$ is the total number of pixels in the frame and $f$ denotes the neighboring frame index. Specifically, $\Delta i = M_{f \rightarrow k}(i)$ indicates that the pixel on the position $i$ of the $f^{th}$ frame moves to the position $i+\Delta i$ of the $k^{th}$ frame.
The estimated optical flow map $M_{f \rightarrow k}$ are applied to the $f^{th}$ frame prediction of the student net $\hat{\boldsymbol y}_{f}^S $ to obtain the warped probability map $\hat{\boldsymbol y}_{f\rightarrow k}^S  $ via resampling operation, i.e. $\hat{\boldsymbol y}^S_{f \rightarrow k} = h(M_{f \rightarrow k}, \hat{\boldsymbol y}^{S}_f)$. The warped probability map serves as temporal constraints for the target frame prediction of student net.

However, error accumulates due to imprecise optical flow estimation and warping noise, which harms the segmentation performance significantly. Therefore, as Figure~\ref{fig:MFD}(b) shows, we raise a well-designed consistency matrix $W_f$ to estimate the quality of the warped probability map $\hat{\boldsymbol y}^{S}_{f\rightarrow k}$ by introducing teacher net prediction $\hat{\boldsymbol y}^{T}_{k}$, i.e. $W_f = \mathrm{exp}(|\hat{\boldsymbol y}^{S}_{f\rightarrow k} - \hat{\boldsymbol y}^{T}_{k}|)$. The teacher net is pre-trained in advance (see Figure~\ref{fig:framework}(a)) and its parameters are fixed in the MFD. Note that the teacher net here is only adopted in the consistency matrix construction instead of directly providing soft knowledge to the student net, which is different from the vanilla knowledge distillation. The consistency matrix adaptively decreases the weight of unreliable region in the warped mask $\hat{\boldsymbol y}^{S}_{f\rightarrow k}$.
For the frame-pair input (i.e. the target frame ${\boldsymbol x_k}$ and the neighboring frame ${\boldsymbol x_f}$), the objective for MFD training scheme is formulated as Eq.~(\ref{eq.mfd}):
\begin{equation}
\label{eq.mfd}
\small
\mathcal{L}_{MFD} ({\boldsymbol x_f}, {\boldsymbol x_k}) =  W_{f} \odot|| \hat{\boldsymbol y}^{S}_{f\rightarrow k} - \hat{\boldsymbol y}_{k}^S ||_{2}^{2},
\end{equation}
where $\odot$ denotes element-wise multiplication. With the proposed MFD training strategy, the memory flow from abundant unlabeled neighboring frames is effectively propagated to the target frame, which stimulates the optimization of the VSS model.

\noindent{\textbf{KD Scheme.}} To alleviate the inference latency issue, we adopt the knowledge distillation strategy to train the compact student net. Note that the teacher net is pre-trained with click annotations under the weakly training scheme in advance (see Figure~\ref{fig:framework}(a)). We train the compact student net with soft knowledge from the teacher net using both the labeled frames and the abundant unlabeled frames. The goal of the student net is to align the class probability of each pixel in every corresponding mask generated by the teacher net. Based on the single-frame input ${\boldsymbol x_f}$, the loss for KD is as follow:
\begin{equation}
\small
\label{eq.MKD}
\begin{split}
\mathcal{L}_{KD}({\boldsymbol x_f})= \mathrm{KL}(\hat{\boldsymbol y}^{S}_{f}||\hat{\boldsymbol y}^{T}_{f}),
\end{split}
\end{equation}
where $\hat{\boldsymbol y}^{S}_{f}$ and $\hat{\boldsymbol y}^{T}_{f} \in R^{ C \times H \times W}$ represent the $f^{th}$ frame prediction maps of the student net and the teacher net, $C$ is the number of class, $H$ and $W$ is the height and width of the prediction maps, and KL denotes Kullback-Leibler divergence.


\noindent{\textbf{Overall Optimization Process.}} The overall optimization process for student net consists of the above three training schemes. Note that only the target frame ${\boldsymbol x_k}$ has click annotation and the $n-1$ neighboring frames $\{\boldsymbol{x}_f\}_{f=k-n+1}^{k-1}$ remain unlabeled. The optimization goal for the training phase is to minimize~Eq. (\ref{eq.total}):
\begin{equation}
\hspace{-3mm}
\small
\label{eq.total}
\mathcal{L} = 
\alpha \mathcal{L}_{weakly}({\boldsymbol x_k})+ \beta\sum_{f=k-n+1}^{k-1}\mathcal{L}_{MFD}({\boldsymbol x_f},{\boldsymbol x_k})+\gamma\sum_{f=k-n+1}^{k}\mathcal{L}_{KD}({\boldsymbol x_f}) ,
\end{equation}
where the $\alpha$, $\beta$ and $\gamma$ are modulating factors for the loss weights.



\section{Experiments} 
\subsection{Datasets and Metrics}

\noindent\textbf{Datasets}. We conduct the experiments on the standard benchmarks, i.e. Cityscapes~\cite{Cordts2016} and Camvid~\cite{Brostow2009}, for video semantic segmentation (VSS). Cityscapes consists of 5000 sparsely-labeled video snippets of urban scenes, each of which contains 30 frames and only the \engordnumber{20} frame is finely annotated in pixel level. The images are divided into 2975, 500, 1525 for training, validation, and testing respectively. Only 19 semantic classes and 1 void class are used for training. As for the Camvid, it contains 4 videos that are annotated at 1 Hz, with 367 images for training, 100 for validation, and 233 for testing.

\noindent\textbf{Metrics}. We apply mean Intersection-over-Union (mIoU)~\cite{long2015fully} and mean Pixel Accuracy (mPA) as the accuracy metrics. Besides, we report the parameters (\#Params) and Frames Per Second (FPS) to evaluate the model efficiency.

\subsection{Implementation Details}

\noindent\textbf{Click Annotations.} In our experiments, we only use click annotations instead of per-pixel annotations for training. Referring to the click generation strategy in~\cite{bearman2016whats} and ~\cite{obukhov2019gated}, we first divide the classes into instance classes (e.g. car and person), and non-instance classes (e.g. building, and road). Concerning instance classes, instance segmentation masks are used to distinguish different instances. In terms of non-instance classes, 8-connectivity component labeling is applied and components with less than 512 pixels are discarded. For the remaining non-instance objects and all the instance objects, we select the centroid point of the semantic object as the click annotation if the point is within the object. Otherwise, we randomly sample one point within the object as the click annotation. It is noteworthy that the click annotation can be an arbitrary point within the semantic object in practice.

\noindent\textbf{Network Architectures.} We apply two popular semantic segmentation networks, i.e. DeeplabV3+~\cite{chen2018encoder} and PSPNet~\cite{Zhao2017Pyramid} to perform the experiments. For the teacher net, ResNet101~\cite{he2016deep} is used as backbone. With respect to compact student net, we adopt ResNet18 and MobileNetV2~\cite{mobilenetv2}. Besides, We employ FlowNetV2~\cite{ilg2017flownet} to estimate the optical flow between two frames.

\noindent\textbf{Experimental Setup.}
The experiments are performed under a single Nvidia Tesla V100 GPU, and Intel(R) Xeon(R) Platinum 8168 CPU @ 2.70GHz. During the training stage, common practices are applied, i.e. backbone pre-training with ImageNet~\cite{deng2009imagenet}, Xavier initialization~\cite{glorot2010understanding}, data augmentations including random crop, random scale in [0.5, 2.0], random horizontal flip, random mirror, etc. As for the Cityscapes dataset, we train the model using stochastic gradient descent (SGD) with initial learning rate 7$e^{-3}$, polynomial learning rate decay scheduler power 0.9, momentum 0.9, weight decay 1$e^{-4}$, batch size 4, and crop size 768 $\times$ 768 for 120 epochs. For the Camvid dataset, we conduct experiments with initial learning rate 1$e^{-4}$, batch size 8, and crop size 360 $\times$ 360 for 200 epochs. Besides, WeClick introduces several hyper-parameters, including memory flow direction, input frame quantity, and sampling policy. `Memory flow direction', is the direction of memory flow conveyed to the target frame, e.g. the \engordnumber{20} frame in Cityscapes, during training. `Input frame quantity', represents the number of frames to feed in the WeClick training pipeline. In other words, input frame quantity $n$ indicates 1 target frame and $n-1$ neighboring frames. `Sampling policy', is the input frames selection strategy during training. We conduct extensive ablation studies in Section~\ref{ablation_study} to show the impact of these hyper-parameters. We set $\lambda=0.1$ in Eq.~(\ref{eq.weak}), following the setting in~\cite{obukhov2019gated}. To balance the loss weights of Eq.~(\ref{eq.total}), we set $\alpha=1$, $\beta=1$, $\gamma=1$ by simple attempts. For the reliability of the results, we conduct the experiments for 5 times and report the average value.

\subsection{Ablation Studies}
\label{ablation_study}

\noindent\textbf{Effectiveness of the Proposed Training Schemes.} To validate the effectiveness of the training schemes, we conduct extensive experiments on DeeplabV3+ with ResNet18 and MobileNetV2. Due to the great regularization capability of the proposed WeClick pipeline, the model accuracy increases dramatically. As Table~\ref{tab:1} shows, the mIoU gains of ResNet18 and MobileNetV2 reach 5.42\% and 5.43\% respectively. It is noteworthy that the compact student net with ResNet18 achieves 58.30\% in terms of mIoU on Cityscapes validation set, nearly surpassing the teacher net with ResNet101, which indicates the proposed method significantly boosts the performance of the lightweight networks in the click-based VSS.

\begin{table}[t]
\renewcommand{\arraystretch}{1.2}
\center
\tabcolsep 0.03in \begin{tabular}{ccccclc}
\hline
   & Backbone & Weakly & KD &  MFD  &  mIoU (\%) & mPA (\%)\\
\hline
Teacher & ResNet101 & \Checkmark   &         &           &  58.35  &  67.24 \\
\hline
\multirow{3}{*}{Student} & ResNet18  & \Checkmark   &            &              &  52.88  & 62.21   \\

& ResNet18    & \Checkmark   & \Checkmark &              &  57.49$^{\uparrow4.61}$  &  65.10 \\
& ResNet18    & \Checkmark   & \Checkmark & \Checkmark   &  \textbf{58.30$^{\uparrow5.42}$}  &  66.39 \\


\hline

\multirow{3}{*}{Student}& MobileNetV2 & \Checkmark   &            &              &  51.99  & 62.77  \\
& MobileNetV2 & \Checkmark   & \Checkmark &              &      56.26$^{\uparrow4.27}$  &  66.14 \\
& MobileNetV2 & \Checkmark   & \Checkmark & \Checkmark   &  \textbf{57.42$^{\uparrow5.43}$}  & 66.89  \\

\hline
\end{tabular}
\vspace{4mm}
\caption{Ablation studies of different training schemes in WeClick. With the proposed training schemes, student nets, i.e. ResNet18 and MobileNetV2, obtain large performance gains in terms of both mIoU and mPA. Note that the ResNet18 and MobileNetV2 results are both under the best hyper-parameter settings.}
\label{tab:1}
\end{table}
\begin{table}[t]
\renewcommand{\arraystretch}{1.2}
\center
\tabcolsep 0.01in \begin{tabular}{ccccc}
\hline
Backbone & MFD Direction   & Input Frames    &   mIoU (\%)    & mPA (\%)\\
\hline

\multirow{3}{*}{ResNet18}  & Forward & \engordnumber{18}, \engordnumber{19}, \textbf{\engordnumber{20}}   & \textbf{58.03}  &  65.99   \\
&    Bi-Direction & \engordnumber{19}, \textbf{\engordnumber{20}}, \engordnumber{21}     & 57.56  &  66.16   \\
&    Backward  & \textbf{\engordnumber{20}}, \engordnumber{21}, \engordnumber{22}  & 57.24  &  65.50   \\
\hline

\multirow{3}{*}{MobileNetV2} &    Forward  & \engordnumber{18}, \engordnumber{19}, \textbf{\engordnumber{20}}   & \textbf{57.21}  &  66.66    \\
&    Bi-Direction   & \engordnumber{19}, \textbf{\engordnumber{20}}, \engordnumber{21}    & 57.09  &  66.12   \\
&    Backward  & \textbf{\engordnumber{20}}, \engordnumber{21}, \engordnumber{22}   & 56.94  & 65.51   \\
\hline
\end{tabular}
\vspace{4mm}
\caption{Comparison on different MFD directions in the student nets DeeplabV3+ with ResNet18 and MobileNetV2. The forward MFD is superior to both bi-direction and backward MFD in terms of mIoU.}
\label{tab:2}
\end{table}

\begin{figure}[t]
\center

\includegraphics[width=1\linewidth]{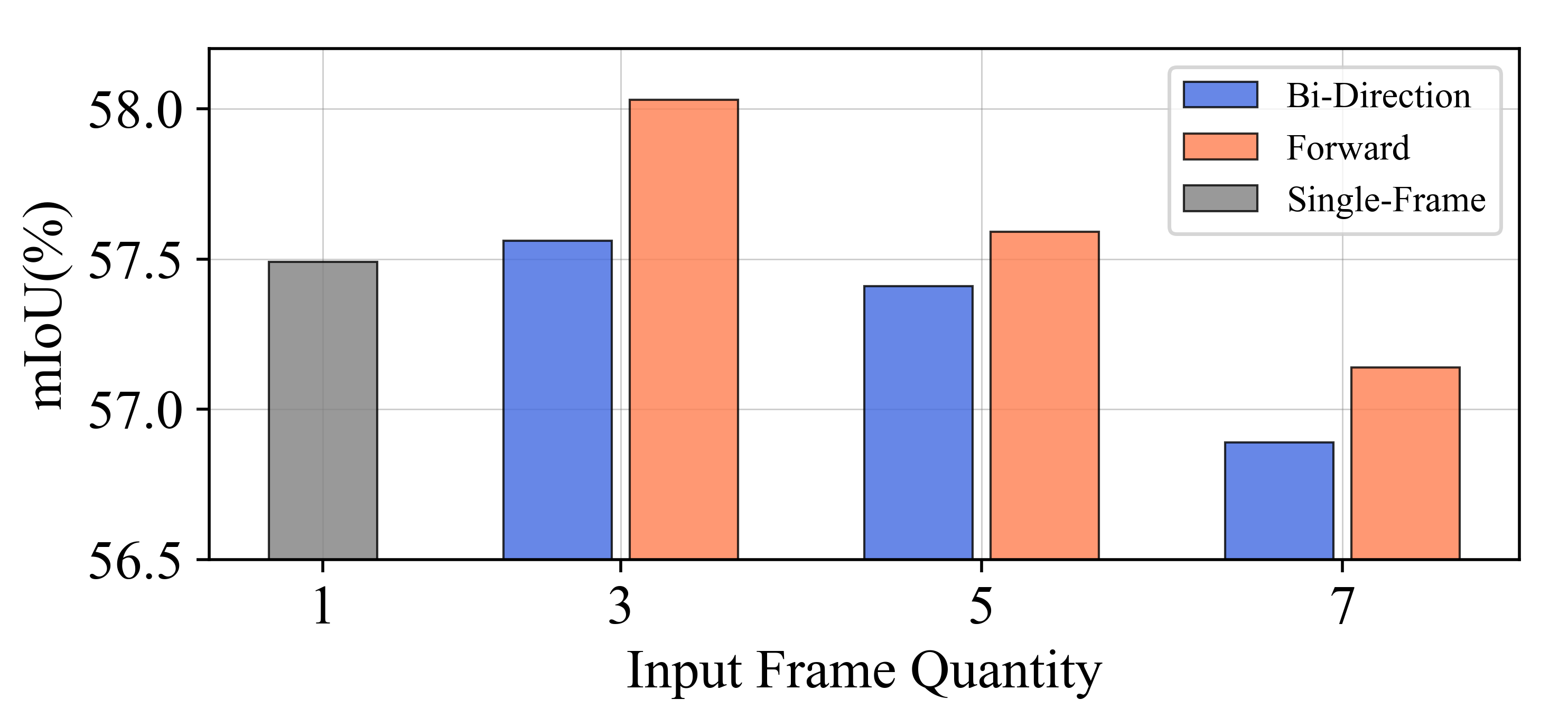}
\caption{Comparison on different input frame quantities in student net DeeplabV3+ with ResNet18 under forward and bi-direction MFD settings. Note that we conduct a single-frame experiment without MFD to act as the baseline. The model trained with 3 input frames performs best.}
\label{fig:frameNum}
\end{figure}

\begin{table}[t]
\renewcommand{\arraystretch}{1.2}
\center
\tabcolsep 0.015in \begin{tabular}{ccccc}
\hline
\multirow{2}{*}{Sampling Policy} & \multicolumn{2}{c}{Forward} &\multicolumn{2}{c}{Bi-Direction} \\
\cline{2-3}\cline{4-5}
         & mIoU (\%) & mPA (\%) & mIoU (\%) & mPA (\%) \\
\hline

 Fixed Frame-Interval 1       &   58.03 & 65.99     &  57.56 & 66.16 \\
 Fixed Frame-Interval 4       &   57.94 & 66.03     &  57.43 & 66.09 \\
Fixed Frame-Interval 7       &   57.83 & 65.50     &  57.41 & 65.69 \\
  \hline
 Random Sampling  &   \textbf{58.30} & 66.39     &  \textbf{57.74} & 66.43  \\

\hline
\end{tabular}
\vspace{4mm}
\caption{Comparison on different sampling policies in the student net DeeplabV3+ with ResNet18 on 3 input frames (including the target frame). Larger sampling intervals are not conducive to improving performance, while random sampling policy achieves performance gain.}
\label{tab:4}
\end{table}
\begin{table}[t]
\renewcommand{\arraystretch}{1.2}
\center
\tabcolsep 0.02in \begin{tabular}{cccccc}
\hline

\hline
Student  & Teacher & mIoU (\%) & mPA (\%) &  \#Params (M) & FPS\\
\hline
DL-R101    & None    & 58.35   &   67.24  &  59.34  & 8.41   \\
\hdashline[3pt/5pt]
DL-R18     & None    & 52.88   &   62.21  &   16.61 &  22.06 \\
DL-R18     & DL-R101 & \textbf{58.30$^{\uparrow5.42}$}  
                               &   66.39  &   16.61 &  22.06 \\
\hdashline[3pt/5pt]
DL-Mob     & None    & 51.99   &   62.77  &  13.35  & 25.97  \\
DL-Mob     & DL-R101 & \textbf{57.42$^{\uparrow5.43}$}   
                               &   66.89  &  13.35 &  25.97 \\
\hline
\hline

\hline
Student  & Teacher & mIoU (\%) & mPA (\%) &  \#Params (M) & FPS\\
\hline
PSP-R101    & None   & 54.42   &   63.09  &  69.31  &  3.42 \\
\hdashline[3pt/5pt]
PSP-R18     & None   & 50.53   &   59.86  &  13.80  & 15.63  \\
PSP-R18     & PSP-R101       & \textbf{55.25$^{\uparrow4.72}$}   &   63.20  &   13.80  &  15.63 \\
\hdashline[3pt/5pt]
PSP-Mob   & None     & 42.20   &   54.32  &  10.31  & 52.23  \\
PSP-Mob   & PSP-R101       & \textbf{52.44$^{\uparrow10.24}$}   &   62.71 &10.31   & 52.23  \\
\hline

\end{tabular}
\vspace{4mm}
\caption{Accuracy (mIoU/mPA) and inference speed (FPS) on Cityscapes validation set. For short notation, DL-R101, DL-R18, and DL-Mob denote DeeplabV3+ with ResNet101, ResNet18, and MobileNetV2 as backbones. PSP-R101, PSP-R18, and PSP-Mob represent PSPNet with ResNet101, ResNet18, and MobileNetV2 as backbones, respectively. Our proposed WeClick improves both the accuracy and inference speed substantially than baseline.}
\label{tab:5}
\end{table}

\noindent\textbf{Impact of the MFD Direction.}
In WeClick, MFD propagates memory flow from preceding frames to the target frame by default, which is denoted as forward MFD. We extend the direction of MFD to three cases, including forward, backward, and bi-direction. In this ablation study, we set the input frame quantity $n$ to 3 (i.e. including 1 target frame and 2 neighboring frames) and select the neighboring frames with the fixed frame-interval 1. Note that we consider the \engordnumber{20} frame in each video snippet as the target frame in Cityscapes. Therefore, backward MFD indicates information transfers from subsequent frames (i.e. \engordnumber{21} and \engordnumber{22} frames) to the target frame, while bi-direction MFD conveys information from both preceding and subsequent frames (i.e. \engordnumber{19} and \engordnumber{21} frames) to the target frame. Table~\ref{tab:2} demonstrates that forward MFD outperforms the other two cases in terms of mIoU on student nets DeeplabV3+ with ResNet18 and MobileNetV2 as backbones.

\noindent\textbf{Impact of the Input Frame Quantity.}
In this section, we investigate the impact of input frame quantity $n$ (i.e. including 1 target frame and $n-1$ neighboring frames), in student net DeeplabV3+ with ResNet18 under both forward and bi-direction MFD settings with the fixed frame-interval 1. Note that we conduct a single-frame experiment without the MFD scheme to act as the baseline. As Figure~\ref{fig:frameNum} illustrates, the model trained with 3 input frames outperforms baseline, which indicates the effectiveness of the proposed MFD strategy. 
However, the performance degrades consistently as the quantity of input frames increases in both forward and bi-direction MFD settings. It can be explained that the gain brought by additional training data and temporal regularization from MFD is suppressed by accumulated optical flow warping error from distant frames, which harms the generalization capability of the student net.


\begin{figure*}[t]
\center
\vspace{-6mm}
\includegraphics[width=0.8\linewidth]{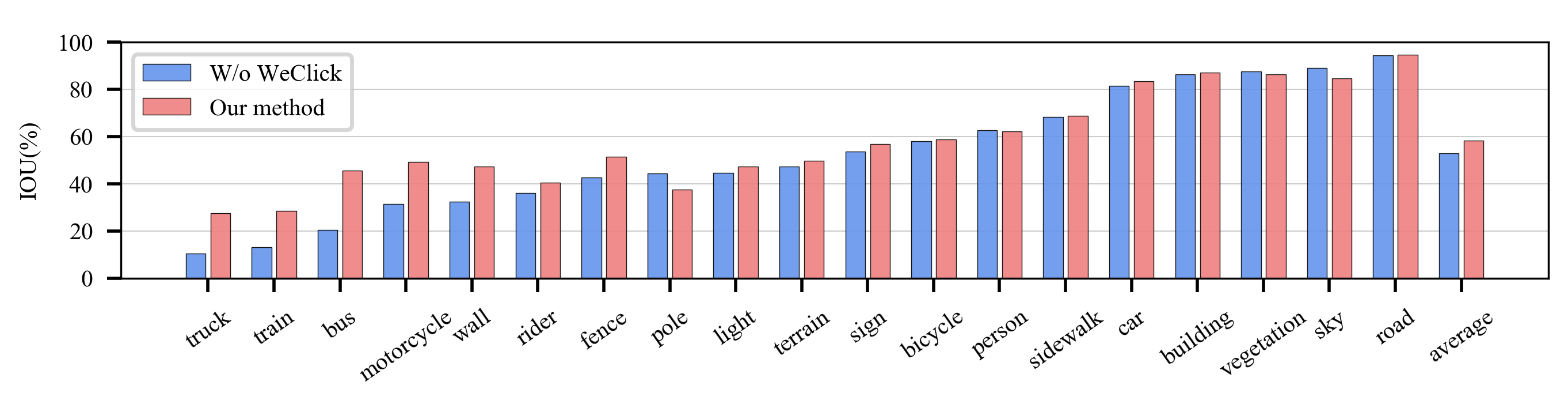}
\caption{Comparison between WeClick and baseline in per-class IoU on DeeplabV3+ with ResNet18. WeClick improves the performance for instance class, such as truck, bus, train, etc, which benefits from the memory flow introduced by MFD.}
\label{fig:cIoU}
\end{figure*}

\begin{table*}[t]
\renewcommand{\arraystretch}{1.2}
\center
\begin{tabular}{cccccccc}
\hline

\hline
\multirow{2}{*}{Method}  & \multirow{2}{*}{Task} & \multirow{2}{*}{Supervision}   &\multirow{2}{*}{\#Params(M)} 
                         & \multicolumn{2}{c}{Cityscapes} & \multicolumn{2}{c}{Camvid}\\
\cline{5-8}
        &            &    & &  mIoU (\%)   & FPS    & mIoU (\%)  & FPS \\
\hline

\hline
Bringing~\cite{sadat2017bringing} 
&  VSS      &   image-level  &    --      &  24.9     &    -   & - &   --  \\
ADVENT~\cite{Vu2019ADVENT} 
&  ISS        &   UDA          &    --      &  48.0     &    7.1   &  --  &   --  \\

Gated CRF (DeeplabV3+ with ResNet34)~\cite{obukhov2019gated} 
&  ISS        &   click-level  &    26.72   &  57.43     &   17.3     &    46.9  & 99.1    \\
Gated CRF (DeeplabV3+ with ResNet18)~\cite{obukhov2019gated} 
&  ISS        &   click-level  &    16.61   &  52.88     &   22.1    &    46.2  & 128.3   \\
\hline
Our WeClick (DeeplabV3+ with ResNet18)
&  VSS        &  click-level  &     16.61   & 58.30  &  22.1  &  48.7  &   128.3   \\
Our WeClick (DeeplabV3+ with MobileNetV2)
&  VSS        &  click-level  &     13.35   & 57.42  & 26.0   &   --   &     --    \\
Our WeClick (PSPNet with ResNet18)
&  VSS        &  click-level  &     13.80   & 55.25  & 15.6   &   --   &     --     \\
Our WeClick (PSPNet with MobileNetV2)
&  VSS        &  click-level  &     10.31   & 52.44  &  52.2  &   --   &     --      \\

\hline
\end{tabular}
\vspace{4mm}
\caption{Comparison with weakly-supervised learning VSS methods on Cityscapes and Camvid. VSS and ISS represent video semantic segmentation and image semantic segmentation, respectively. The label types include image-level, click-level, and unsupervised domain adaptation (UDA). The results shows WeClick (DeeplabV3+ with ResNet18) outperforms other methods consistently by a large margin and WeClick with MobileNetV2 reaches a real-time inference speed.}
\label{tab:6}
\end{table*}

\noindent\textbf{Impact of the Sampling Policy.}
By default, under 3 input frames setting, we select input frames from the consecutive neighbors of the target frame with the fixed frame-interval 1. In this experiment, we investigate the impact of different sampling policies on student net with 3 input frames, including the fixed frame-intervals (i.e. 1, 4, and 7), and random sampling strategy in a predefined range (i.e. from \engordnumber{16} to \engordnumber{19} in the preceding frames and from \engordnumber{21} to \engordnumber{24} in the subsequent frames). For instance, in the setting of forward MFD with frame-interval 4, the selected input frames are \engordnumber{12}, \engordnumber{16} and \engordnumber{20}. Table~\ref{tab:4} reveals that, on the one hand, with a larger fixed sampling frame-interval, the performance declines because of the optical flow estimation noise between two distant frames. On the other hand, with the random sampling policy, the mIoU is improved slightly compared to the best results in the fixed frame-interval settings, i.e. frame-interval 1. It can be explained that random sampling policy sees more unlabeled video data, and capture both short-term and long-term temporal information when training.

\noindent\textbf{Impact of the Teacher Net and Student Net.}
In this section, we show the generalization of our WeClick in different teacher-student settings. Table~\ref{tab:5} represents that our proposed method consistently improves the performance of student nets without compromising efficiency during inference. Specifically, student net PSPNet with ResNet18 surpasses teacher net and outperforms baseline by 4.72\% mIoU. Moreover, MobileNetV2 in both DeeplabV2 and PSPNet reach real-time execution under Nvidia Tesla V100 with only slight accuracy compromise compared with the teacher net, which indicates the effectiveness and generalization of WeClick.



\subsection{Comparison with the State-of-the-art Methods}

In this section, we compare our proposed WeClick with state-of-the-art weakly-supervised methods, whose results are from their papers. Table~\ref{tab:6} demonstrates that our method outperforms other state-of-the-art weakly-supervised methods, including image-level, click-level and UDA weakly-supervised algorithms, by a large margin. For example, WeClick (DeeplabV3+ with ResNet18) improves the segmentation performance than ~\cite{sadat2017bringing} and~\cite{Vu2019ADVENT} by 33.4\% mIoU and 10.3\% mIoU, respectively.
It is noteworthy that WeClick is superior to state-of-the-art click-based image semantic segmentation method Gated CRF~\cite{obukhov2019gated} in both accuracy and inference speed. 

Besides, Figure \ref{fig:cIoU} shows the per-class IoU of the methods with and without WeClick. Our method significantly improves the performance of instance classes, such as truck, bus, train, etc. It can be explained that the memory flow enables the object classes with significantly changing motion to perform better.


We further conduct additional experiments in DeeplabV3+ with ResNet18 on the CamVid dataset to validate the generalization of the WeClick across different datasets. As Table~\ref{tab:6} shows, our proposed WeClick stimulates the capability of the model and outperforms the baseline by 2.5\% mIoU. Our method can achieve 128.3 fps with a $352\times480$ resolution on a single Nvidia Tesla V100. Consistent mIoU improvements on both the Cityscapes and the Camvid verify both the generalization and effectiveness of our proposed WeClick pipeline for click-based VSS.

\section{Conclusions}
In this work, we make the first attempt to propose an effective weakly-supervised video semantic segmentation pipeline with click annotations, called WeClick, to save tedious annotating effort by segmenting an instance of the semantic class with only a single click during the training phase. To boost the performance, we design a novel memory flow knowledge distillation training scheme, which utilizes temporal information in the unlabeled video frames. WeClick learns compact video semantic segmentation models with the low-cost click annotations in the training period yet achieves real-time and accurate models during the inference phase. Extensive experiments on the Cityscapes and Camvid show that the compact models with the proposed training schemes outperform state-of-the-art methods by a large margin, which demonstrates the effectiveness and generalization of WeClick. We hope our work could provide insights for researchers of this field to design novel schemes for weakly-supervised video semantic segmentation.


\clearpage
\bibliographystyle{ACM-Reference-Format}
\balance
\bibliography{ms}

\clearpage


\end{document}